# Boltzmann Machine Learning with the Latent Maximum Entropy Principle

**Shaojun Wang**
University of Toronto
Toronto, Canada

**Dale Schuurmans**
University of Waterloo
Waterloo, Canada

**Fuchun Peng**
University of Waterloo
Waterloo, Canada

**Yunxin Zhao**
University of Missouri
Columbia, USA

## Abstract

We present a new statistical learning paradigm for Boltzmann machines based on a new inference principle we have proposed: the latent maximum entropy principle (LME). LME is different both from Jaynes' maximum entropy principle and from standard maximum likelihood estimation. We demonstrate the LME principle by deriving new algorithms for Boltzmann machine parameter estimation, and show how a robust and rapidly convergent new variant of the EM algorithm can be developed. Our experiments show that estimation based on LME generally yields better results than maximum likelihood estimation when inferring models from small amounts of data.

## 1 Introduction

Boltzmann machines are probabilistic networks of binary valued random variables that have wide application in areas of pattern recognition and combinatorial optimization (Aarts and Korst 1989). A Boltzmann machine can be represented as an undirected graphical model whose associated probability distribution has a simple quadratic energy function, and hence has the form of a Boltzmann-Gibbs distribution.

Typically, the random variables are divided into a set of visible variables, whose states are clamped at observed data values, and a separate set of hidden variables, whose values are unobserved. Inference, therefore, usually consists of calculating the conditional probability of a configuration over the hidden variables, or finding a most likely configuration of the hidden variables, given observed values for the visible variables. Inference in Boltzmann machines is known to be hard, because this generally involves summing or maximizing over an exponential number of possible configurations of the hidden variables (Ackley et al. 1985, Welling and Teh 2003).

In this paper we will focus on the key problem of estimating the parameters of a Boltzmann machine from data. There is a surprisingly simple algorithm (Ackley et al. 1985) for performing maximum likelihood estimation of the weights of a Boltzmann machine—or equivalently, to minimize the KL divergence between the empirical data distribution and the implied Boltzmann distribution. This classical algorithm is based on a direct gradient ascent approach where one calculates (or estimates) the derivative of the likelihood function with respect to the model parameters. The simplicity and locality of this gradient ascent algorithm has attracted a great deal of interest.

However, there are two significant problems with the standard Boltzmann machine training algorithm that we address in this paper: First, the convergence rate of the standard algorithm is extremely slow. It is well known that gradient ascent converges only linearly, with a rate that depends critically on the condition of the Hessian matrix of the likelihood function. In order to achieve faster convergence one generally has to adjust the step size. In practice, gradient ascent methods often select step sizes by performing an explicit line search in the gradient direction, which can lead to non-trivial overhead.

Second, it is known that the true distribution is multimodal, and the likelihood function has multiple local maxima. This raises the question of which local maximizer to choose as the final estimate. Fisher's classical maximum likelihood estimation (MLE) principle states that the desired estimate corresponds to the global maximizer of the likelihood function. However, in practice it is often observed that MLE leads to over-fitting (poor generalization) particularly when faced with limited training data.

To address both of the above problems in the context of Boltzmann machines, we have recently proposed a new statistical machine learning framework for density estimation and pattern classification, which we refer to as the latent maximum entropy (LME) principle (Wang et al. 2003).[1] Although classical statistics is heavily based on parametric models, such an approach can sometimes be overly restrictive and prevent accurate models from being developed. The alternative principle we propose, LME, is a non-parametric approach based on matching a set of features in the data (i.e. sufficient statistics, weak learners, or basis functions). This technique becomes parametric when we necessarily have to approximate the principle. LME is an extension to Jaynes' maximum entropy (ME) principle that explicitly incorporates latent variables in the formulation, and thereby extends the original principle to cases where data components are missing. The resulting principle is different from both maximum likelihood estimation and standard maximum entropy, but often yields better estimates in the presence of hidden variables and limited training data.

In this paper we demonstrate the LME principle by deriving new algorithms for Boltzmann machine parameter estimation, and show how a robust and rapidly convergent new variant of the EM algorithm can be developed. Our experiments show that estimation based on LME generally yields better results than maximum likelihood estimation, particularly when inferring hidden units from small amounts of data.

The remainder of the paper is organized as follows. First, in Section 2 we introduce the latent maximum entropy principle, and then outline a general training algorithm for this principle in Section 3. Once these preliminaries are in place, we then present the main contribution of this paper—a new training algorithm for Boltzmann machine estimation—in Section 4. Section 5 compares the new parameter optimization procedure to standard procedures, and Section 6 compares the generalization performance of the new estimation principle to standard maximum likelihood. We conclude the paper with a brief discussion in Section 7.

---

[1] A third problem, which we do not directly address in this paper, is that calculating a gradient involves summing over exponentially many configurations of the hidden variables, and thus becomes intractable in large networks. It is therefore usually impractical to perform exact gradient ascent in these models, and some form of approximate inference (Welling and Teh 2003) or Monte Carlo estimation (Ackley et al. 1985) is needed to approximate the gradient. For the purposes of this paper, we simply assume that some sort of reasonable approximation technique is available. See (Southey et al. 2003) for an attempt to improve classical estimators for this problem.

## 2 The latent maximum entropy principle

To formulate the LME principle, let $X \in \mathcal{X}$ be a random variable denoting the complete data, $Y \in \mathcal{Y}$ be the observed incomplete data and $Z \in \mathcal{Z}$ be the missing data. That is, $X = (Y, Z)$. If we let $p(x)$ and $p(y)$ denote the densities of $X$ and $Y$ respectively, and let $p(z|y)$ denote the conditional density of $Z$ given $Y$, then $p(y) = \int_{z \in \mathcal{Z}} p(x) \, \mu(dz)$ where $p(x) = p(y)p(z|y)$.

**LME principle** Given features $f_1, ... f_N$, specifying the properties we would like to match in the data, select a joint probability model $p^*$ from the space of all distributions $\mathcal{P}$ over $\mathcal{X}$ to maximize the joint entropy

$$H(p) = -\int_{x \in \mathcal{X}} p(x) \log p(x) \, \mu(dx) \quad (1)$$

subject to the constraints

$$\int_{x \in \mathcal{X}} f_i(x) \, p(x) \, \mu(dx) = \sum_{y \in \tilde{\mathcal{Y}}} \tilde{p}(y) \int_{z \in \mathcal{Z}} f_i(x) \, p(z|y) \, \mu(dz)$$

$$i = 1...N, \, Y \text{ and } Z \text{ not independent} \quad (2)$$

where $x = (y, z)$. Here $\tilde{p}(y)$ is the empirical distribution over the observed data, and $\tilde{\mathcal{Y}}$ denotes the set of observed $Y$ values. Intuitively, the constraints specify that we require the expectations of $f_i(X)$ in the complete model to match their empirical expectations on the incomplete data $Y$, taking into account the structure of the dependence of the unobserved component $Z$ on $Y$.

Unfortunately, there is no simple solution for $p^*$ in (1,2). However, a good approximation can be obtained by restricting the model to have an exponential form

$$p_\lambda(x) = \Phi_\lambda^{-1} \exp\left(\sum_{i=1}^{N} \lambda_i f_i(x)\right)$$

where $\Phi_\lambda = \int_{x \in \mathcal{X}} \exp\left(\sum_{i=1}^{N} \lambda_i f_i(x)\right) \mu(dx)$ is a normalizing constant that ensures $\int_{x \in \mathcal{X}} p_\lambda(x) \, \mu(dx) = 1$. This restriction provides a free parameter $\lambda_i$ for each feature function $f_i$. By adopting such a "log-linear" restriction, it turns out that we can formulate a practical algorithm for approximately satisfying the LME principle.

## 3 A general training algorithm for log-linear models

To derive a practical training algorithm for log-linear models, we exploit the following intimate connection between LME and maximum likelihood estimation (MLE).

**Theorem 1** *Under the log-linear assumption, maximizing the likelihood of log-linear models on incomplete data is equivalent to satisfying the feasibility constraints of the LME principle. That is, the only distinction between MLE and LME in log-linear models is that, among local maxima (feasible solutions), LME selects the model with the maximum entropy, whereas MLE selects the model with the maximum likelihood (Wang et al. 2003).*

This connection allows us to exploit an EM algorithm (Dempster et al. 1977) to find *feasible* solutions to the LME principle. It is important to emphasize, however, that EM will only find alternative feasible solutions, while the LME and MLE principles will differ markedly in the feasible solutions they prefer. We illustrate this distinction below.

To formulate an EM algorithm for learning log-linear models, first decompose the log-likelihood function $L(\lambda)$ into

$$L(\lambda) = \sum_{y \in \tilde{\mathcal{Y}}} \tilde{p}(y) \log p_\lambda(y) = Q(\lambda, \lambda') + H(\lambda, \lambda')$$

where $Q(\lambda, \lambda') = \sum_{y \in \tilde{\mathcal{Y}}} \tilde{p}(y) \int_{z \in \mathcal{Z}} p_{\lambda'}(z|y) \log p_\lambda(x) \mu(dz)$, $H(\lambda, \lambda') = -\sum_{y \in \tilde{\mathcal{Y}}} \tilde{p}(y) \int_{z \in \mathcal{Z}} p_{\lambda'}(z|y) \log p_\lambda(z|y) \mu(dz)$. This is a standard decomposition used for deriving EM. For log-linear models, in particular, we have

$$Q(\lambda, \lambda^{(j)}) = -\log(\Phi_\lambda) + \quad (3)$$
$$\sum_{i=1}^{N} \lambda_i \left( \sum_{y \in \tilde{\mathcal{Y}}} \tilde{p}(y) \int_{z \in \mathcal{Z}} f_i(x) \, p_{\lambda^{(j)}}(z|y) \, \mu(dz) \right)$$

Interestingly, it turns out that maximizing $Q(\lambda, \lambda^{(j)})$ as a function of $\lambda$ for fixed $\lambda^{(j)}$ (the M step) is equivalent to solving another constrained optimization problem corresponding to a maximum entropy principle; but a much simpler one than before (Wang et al. 2003).

**Lemma 1** *Maximizing $Q(\lambda, \lambda^{(j)})$ as a function of $\lambda$ for fixed $\lambda^{(j)}$ is equivalent to solving*

$$\max_\lambda \quad H(p_\lambda) = -\int_{x \in \mathcal{X}} p_\lambda(x) \log p_\lambda(x) \mu(dx) \quad (4)$$

$$\text{subject to} \quad \int_{x \in \mathcal{X}} f_i(x) \, p_\lambda(x) \, \mu(dx) = \quad (5)$$

$$\sum_{y \in \tilde{\mathcal{Y}}} \tilde{p}(y) \int_{z \in \mathcal{Z}} f_i(x) \, p_{\lambda^{(j)}}(z|y) \, \mu(dz), \quad i = 1...N$$

It is critical to realize that the new constrained optimization problem in Lemma 1 is much easier than maximizing (1) subject to (2) for log-linear models, because the right hand side of the constraints (5) no longer depends on $\lambda$ but rather on the fixed constants from the previous iteration $\lambda^{(j)}$. This means that maximizing (4) subject to (5) with respect to $\lambda$ is now a convex optimization problem with linear constraints. The generalized iterative scaling algorithm (GIS) (Darroch et al. 1972) or improved iterative scaling algorithm (IIS) (Della Pietra et al. 1997) can be used to maximize $Q(\lambda, \lambda^{(j)})$ very efficiently.

From these observations, we can recover feasible log-linear models by using an algorithm that combines EM with nested iterative scaling to calculate the M step.

**EM-IS algorithm:**

**E step**: Given $\lambda^{(j)}$, for each feature $f_i$, $i = 1, ..., N$, calculate its current expectation $\eta_i^{(j)}$ with respect to $\lambda^{(j)}$ by:

$$\eta_i^{(j)} = \sum_{y \in \tilde{\mathcal{Y}}} \tilde{p}(y) \int_{z \in \mathcal{Z}} f_i(x) \, p_{\lambda^{(j)}}(z|y) \, \mu(dz) \quad (6)$$

**M step**: Perform $S$ iterations of full parallel update of parameter values $\lambda_1, ..., \lambda_N$ either by GIS or IIS as follows. Each update is given by

$$\lambda_i^{(j+s/S)} = \lambda_i^{(j+(s-1)/S)} + \gamma_i \quad (7)$$

such that $\gamma_i$ satisfies

$$\int_{x \in \mathcal{X}} f_i(x) e^{\gamma_i f(x)} p_{\lambda^{(j+(s-1)/S)}}(x) \mu(dx) = \eta_i^{(j)} \quad (8)$$

where $f(x) = \sum_{k=1}^{N} f_k(x)$ and $s = 1, ..., S$. ∎

Provided that the E and M steps can both be computed, EM-IS can be shown to converge to a local maximum in likelihood for log-linear models, and hence is guaranteed to yield feasible solutions to the LME principle.

**Theorem 2** *The EM-IS algorithm monotonically increases the likelihood function $L(\lambda)$, and all limit points of any EM-IS sequence $\{\lambda^{(j+s/S)}, j \geq 0, s = 1..S\}$, belong to the set $\Theta = \{\lambda \in \Re^N : \partial L(\lambda)/\partial \lambda = 0\}$. Therefore, EM-IS asymptotically yields feasible solutions to the LME principle for log-linear models (Wang et al. 2003).*

Thus, EM-IS provides an effective means to find feasible solutions to the LME principle. (We note that Lauritzen (1995) has suggested a similar algorithm, but did not supply a convergence proof. More recently, Riezler (1999) has also proposed an algorithm equivalent to setting $S = 1$ in EM-IS. However, we have found $S > 1$ to be more effective in many cases.)

We can now exploit the EM-IS algorithm to develop a practical approximation to the LME principle.

**ME-EM-IS algorithm:**

*Initialization*: Choose random initial values for $\lambda$.

***EM-IS***: Run EM-IS to convergence, to obtain feasible $\lambda^*$.

***Entropy calculation***: Calculate the entropy of $p_{\lambda^*}$.

***Model selection***: Repeat the above steps several times to produce a set of distinct feasible candidates. Choose the feasible candidate that achieves the highest entropy. ∎

This leads to a new estimation technique that we will compare to standard MLE below. One apparent complication, first, is that we need to calculate the entropies of the candidate models produced by EM-IS. However, it turns out that we do not need to calculate entropies explicitly because one can recover the entropy of *feasible* log-linear models simply as a byproduct of running EM-IS to convergence.

**Corollary 1** *If $\lambda^*$ is feasible, then $Q(\lambda^*, \lambda^*) = -H(p_{\lambda^*})$ and $L(\lambda^*) = -H(p_{\lambda^*}) + H(\lambda^*, \lambda^*)$.*

Therefore, at a feasible solution $\lambda^*$, we have already calculated the entropy, $-Q(\lambda^*, \lambda^*)$, in the M step of EM-IS.

To draw a clear distinction between LME and MLE, assume that the term $H(\lambda^*, \lambda^*)$ from Corollary 1 is constant across different feasible solutions. Then MLE, which maximizes $L(\lambda^*)$, will choose the model that has lowest entropy, whereas LME, which maximizes $H(p_{\lambda^*})$, will chose a model that has least likelihood. (Of course, $H(\lambda^*, \lambda^*)$ will not be constant in practice and the comparison between MLE and LME is not so straightforward, but this example does highlight their difference.) The fact that LME and MLE are different raises the question of which method is the most effective when inferring a model from sample data.

## 4 Application to Boltzmann machine training

Consider a graphical model with $M$ binary nodes taking on values in $\{0,1\}$. Assume that among these nodes there are $J$ observed nodes $Y = (Y_1, ..., Y_J)$, and $L = M - J$ hidden nodes $Z = (Z_1, ..., Z_L)$. Let $X = (Y, Z)$. Thus, $\mathcal{Y} = \{0,1\}^J$, $\mathcal{Z} = \{0,1\}^L$ and $\mathcal{X} = \{0,1\}^{J+L} = \{0,1\}^M$. For this problem, the observed data has the form of a $J$ dimensional vector $y = (y_1, ..., y_J) \in \{0,1\}^J$. Given an observed sequence of $T$ $J$-dimensional vectors $\mathcal{Y} = (y^1, ..., y^T)$, where $y^t \in \{0,1\}^J$ for $t = 1, ..., T$, we attempt to infer a latent maximum entropy model that matches expectations on features defined between every pair of variables in the model. Specifically, we consider the features $f_{k\ell}(x) = y_k y_\ell$, $f_{km}(x) = y_k z_m$, $f_{mn}(x) = z_m z_n$, for $1 \leq k < \ell \leq J$ and $1 \leq m < n \leq L$, where $x = (y, z) = (y_1, ..., y_J, z_1, ..., z_L)$. Note that the features are all binary, and therefore we can represent the structure of the log-linear model by a graph, as shown in Figure 1.

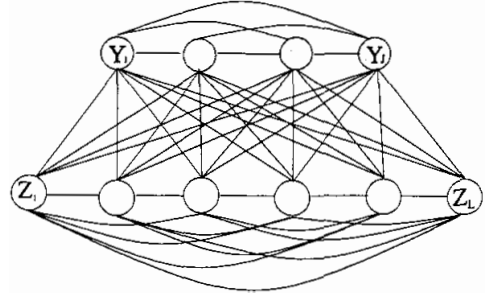

Figure 1: Boltzmann machine model: the nodes $Y$ are observed and the nodes $Z$ are unobserved.

Given a sequence of observed data $\tilde{\mathcal{Y}} = (y^1, ..., y^T)$, we formulate the LME principle as

$$\max_{p(x)} H(X) = H(Y) + H(Z|Y) \quad (9)$$

subject to

$$\sum_{x \in \mathcal{X}} y_k y_\ell \, p(x) = \sum_{y \in \tilde{\mathcal{Y}}} y_k y_\ell \, \tilde{p}(y)$$

$$\sum_{x \in \mathcal{X}} y_k z_m \, p(x) = \sum_{y \in \tilde{\mathcal{Y}}} y_k \, \tilde{p}(y) \sum_{z \in \{0,1\}^L} z_m \, p(z|y)$$

$$\sum_{x \in \mathcal{X}} z_m z_n \, p(x) = \sum_{z \in \{0,1\}^L} z_m z_n \, p(z)$$

for $1 \leq k < \ell \leq J$ and $1 \leq m < n \leq L$

where $x = (y, z) = (y_1, ..., y_J, z_1, ..., z_L)$ and $\tilde{p}(y) = \frac{1}{T}$.

Again we can apply EM-IS to find a feasible log-linear model. To execute the E step, calculate the feature expectations according to (6)

$$\eta_{k,\ell}^{(j)} = \frac{1}{T} \sum_{t=1}^{T} y_k^t y_\ell^t$$

$$\eta_{k,m}^{(j)} = \frac{1}{T} \sum_{t=1}^{T} y_k^t \sum_{z \in \{0,1\}^L} z_m \, p(z|y^t)$$

$$\eta_{m,n}^{(j)} = \sum_{z \in \{0,1\}^L} z_m z_n \, p(z)$$

for $1 \leq k < \ell \leq J$ and $1 \leq m < n \leq L$

To execute the M step we then formulate the simpler maximum entropy problem with linear constraints, as in (4) and (5)

$$\max_{p(x)} H(X) = H(Y) + H(Z|Y) \quad (10)$$

$$\text{subject to} \quad \sum_{x \in \mathcal{X}} y_k y_\ell \, p(x) = \eta_{k,\ell}^{(j)}$$

$$\sum_{x \in \mathcal{X}} y_k z_m \, p(x) = \eta_{k,m}^{(j)}$$

$$\sum_{x \in \mathcal{X}} z_m z_n \, p(x) = \eta_{m,n}^{(j)}$$

for $1 \leq k < \ell \leq J$ and $1 \leq m < n \leq L$

where $x = (y, z) = (y_1, ..., y_J, z_1, ..., z_L)$. In this case, the probability distribution for the complete data model can be written

$$p_\Lambda(x) = p_\Lambda(z, y)$$
$$= \frac{1}{\Phi_\Lambda} e^{\frac{1}{2} y^\top \Lambda_Y y + \frac{1}{2} z^\top \Lambda_Z z + y^\top \Lambda_{YZ} z}$$
$$= \frac{1}{\Phi_\Lambda} e^{\frac{1}{2} x^\top \Lambda x}$$

where $\Lambda = \begin{bmatrix} \Lambda_Y & \Lambda_{YZ} \\ \Lambda_{YZ} & \Lambda_Z \end{bmatrix}$ is the $M \times M$ symmetric matrix of $\lambda$ parameters corresponding to the features over the variable pairs (with the diagonal elements of $\Lambda$ equal to zero), and $\Phi_\Lambda = \sum_{x \in \{0,1\}^M} e^{\frac{1}{2} x^\top \Lambda x}$ is the normalization factor. This graphical model corresponds to a Boltzmann machine (Ackley et al. 1985). To solve for the optimal Lagrange multipliers $\Lambda^{(j)}$ in the M step we once again need to use iterative scaling. Following (7), we iteratively improve $\Lambda^{(j)}$ by adding the update parameters $\gamma^{(j+s/S)}$ that satisfy (8). These can be calculated by by using Newton's method or the bisection method to solve for $\gamma^{(j+s/S)}$ in

$$\sum_{x \in \{0,1\}^M} \frac{1}{\Phi_{\Lambda^{(j+(s-1)/S)}}} y_k y_\ell$$
$$\exp\left(\frac{1}{2} x^\top \left[\Lambda^{(j+(s-1)/S)} + \gamma_{k,\ell}^{(j+s/S)}(\mathbf{1}^\top \mathbf{1} - I_M)\right] x\right) = \eta_{k,\ell}^{(j)}$$

$$\sum_{x \in \{0,1\}^M} \frac{1}{\Phi_{\Lambda^{(j+(s-1)/S)}}} y_k z_m$$
$$\exp\left(\frac{1}{2} x^\top \left[\Lambda^{(j+(s-1)/S)} + \gamma_{k,i}^{(j+s/S)}(\mathbf{1}^\top \mathbf{1} - I_M)\right] x\right) = \eta_{k,m}^{(j)}$$

$$\sum_{x \in \{0,1\}^M} \frac{1}{\Phi_{\Lambda^{(j+(s-1)/S)}}} z_m z_n$$
$$\exp\left(\frac{1}{2} x^\top \left[\Lambda^{(j+(s-1)/S)} + \gamma_{i,j}^{(j+s/S)}(\mathbf{1}^\top \mathbf{1} - I_M)\right] x\right) = \eta_{m,n}^{(j)}$$

for $1 \leq k < \ell \leq J$ and $1 \leq m < n \leq L$

Here **1** is the $M$ dimensional vector with all 1 elements, and $I_M$ is the $M \times M$ identity matrix. The required expectations can be calculated by direct enumeration when $M$ is small, or approximated by Monte Carlo estimation (Ackley et al. 1985), mean field theory (Kappen ad Rodriguez, 1998) or generalized belief propagation (Wainwright et al. 2003, Welling and Teh, 2003, Yeddida et al. 2002) when $M$ is large.

## 5　Comparing EM-IS training to standard EM

To compare EM-IS to standard Boltzmann machine estimation techniques, first consider the derivation of a direct EM approach. In standard EM, given the previous parameters $\Lambda^{(j)}$, one solves for new parameters $\Lambda$ by maximizing the auxiliary $Q$ function with respect to $\Lambda$

$$Q(\Lambda, \Lambda') = \frac{1}{T} \sum_{t=1}^{T} \sum_{z \in \{0,1\}^L} p_{\Lambda'}(z|y^t) \log p_\Lambda(y^t, z)$$

$$= -\log(\Phi_\Lambda) + \frac{1}{2T} \sum_{t=1}^{T} \sum_{z \in \{0,1\}^L} x^\top \Lambda x \, p_{\Lambda'}(z|y^t)$$

Taking derivatives with respect to $\Lambda$ gives

$$\frac{\partial}{\partial \Lambda} Q(\Lambda, \Lambda') \tag{11}$$
$$= -\frac{1}{2} E_{p_\Lambda}[xx^\top] + \frac{1}{2T} \sum_{t=1}^{T} \sum_{z \in \{0,1\}^L} xx^\top p_{\Lambda'}(z|y^t)$$

Apparently there is no closed form solution to the M step and a generalized EM algorithm has to be used in this case. The standard approach is to use a gradient ascent to approximately solve the M step. However, the step size needs to be controlled to ensure a monotonic improvement in $Q$.

EM-IS has distinct advantages over the standard gradient ascent EM approach. First, EM-IS completely avoids the use of tuning parameters while still guaranteeing monotonic improvement. Moreover, we have found that EM-IS converges faster than gradient ascent EM. Figure 2 shows the result of a simple experiment that compares the rate of convergence of M step optimization techniques on a small Boltzmann machine with five visible nodes and three hidden nodes. Comparing EM-IS to the gradient ascent EM algorithm proposed in (Ackley et al. 1985), we find that EM-IS obtains substantially faster convergence. Figure 2 also shows that using several IS iterations in the inner loop, $S = 4$, yields faster convergence than taking a single IS step, $S = 1$ (which corresponds to Riezler's proposed algorithm (Riezler 1999)).

We note that in previous work, Byrne (1992) has proposed a sequential update algorithm for the M step in a Boltzmann machine parameter estimation algorithm. However, to maintain monotonic convergence,

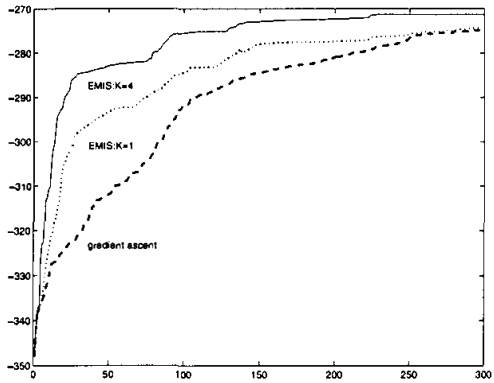

Figure 2: Convergence evaluation: log-likelihood versus iteration, solid curve denotes EM-IS with S=4, dotted curve denotes EM-IS with S=1,and dashed curve denotes gradient ascent.

Byrne's algorithm requires a large number of iterations in the M step to ensure that a maximum is achieved. Otherwise the monotonic convergence property can be violated for the sequential updates he proposes. In our case, EM-IS uses a parallel update procedure that avoids this difficulty. A sequential algorithm that maintains the monotonic convergence property can also be achieved along the lines of (Collins et al. 2002).

## 6 Comparing ME-EM-IS estimation to maximum likelihood

Even assuming that one has an effective algorithm for local parameter optimization, there remains the issue of coping with multiple local maxima. To select from local maxima we propose to use the new latent maximum entropy (LME) technique instead of the classical maximum likelihood (MLE) approach. To ascertain whether LME or MLE yields better estimates when inferring models from sample data that has missing component, we conducted a simple experiment. In particular, we considered a simple eight nodes Boltzmann machine with 5 observable and 3 hidden units as a case study.

The basis for comparison between LME and MLE is to realize that by the discussion in Section 3, any feasible solution to the LME principle corresponds to a locally maximum likelihood model as specified by (11). Therefore, we can implement EM-IS as outlined in Section 3 and generate feasible candidates for the LME and MLE principles simultaneously. From Theorem 1, we know that LME and MLE consider the same set of feasible candidates, except that among feasible solutions, LME selects the model with the highest entropy, whereas MLE selects the model with the highest likelihood. Corollary 1 shows that these are not equivalent.

We are interested in determining which method yields better estimates of various underlying models $p^*$ used to generate the data. We measure the quality of an estimate $p_\Lambda$ by calculating the *cross entropy* from the correct marginal distribution $p^*(y)$ to the estimated marginal distribution $p_\Lambda(y)$ on the *observed* data component $Y$

$$D(p^*(y)\|p_\Lambda(y)) = \int_{y\in\mathcal{Y}} p^*(y) \log \frac{p^*(y)}{p_\Lambda(y)} \mu(dy)$$

The goal is to minimize the cross entropy between the marginal distribution of the estimated model $p_\Lambda$ and the correct marginal $p^*$. A cross entropy of zero is obtained only when $p_\Lambda(y)$ matches $p^*(y)$.

We consider a series of experiments with different models and different sample sizes to test the robustness of both LME and MLE to sparse training data. In particular, we used the following experimental design.

1. Fix a target Boltzmann machine model $p^*(x) = p^*(y, z)$.

2. Generate a sample of observed data $\tilde{\mathcal{Y}} = (y_1, ..., y_T)$ according to $p^*(y)$.

3. Run EM-IS to generate multiple feasible solutions by starting from 100 random initial parameters $\Lambda$.

4. Calculate the entropy and likelihood for each feasible candidate.

5. Select the maximum entropy candidate $p_{LME}$ as the LME estimate, and the maximum likelihood candidate $p_{MLE}$ as the MLE estimate.

6. Calculate the cross entropy from $p^*(y)$ to the marginals $p_{LME}(y)$ and $p_{MLE}(y)$ respectively.

7. Repeat Steps 2 to 6 5 times and compute the average of the respective cross entropies. That is, average the cross entropy over 5 repeated trials for each sample size and each method, in each experiment.

8. Repeat Steps 2 to 7 for different sample sizes $T$.

9. Repeat Steps 1 to 8 for different generative models $p^*(x)$.

In this experiment we generated the data according to a Boltzmann machine with 5 observable and 3 hidden units, and attempted to learn the parameters for a Boltzmann machine that assumed the same architecture.

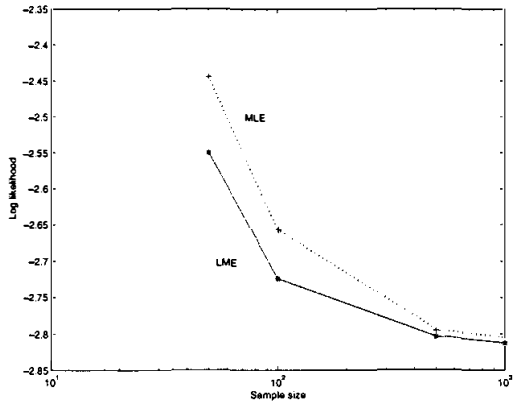

Figure 3: Average log-likelihood of the MLE estimate versus the LME estimates in Experiment 1.

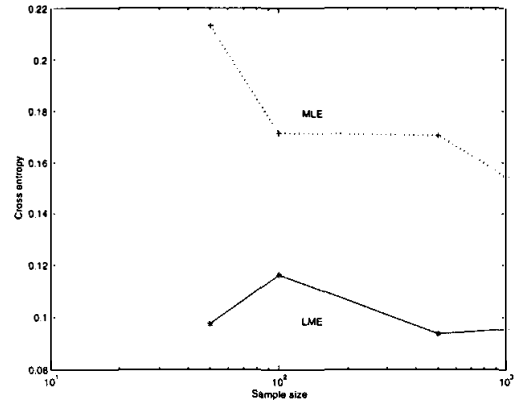

Figure 5: Average cross entropy between the true distribution and the MLE estimate versus the LME estimates in Experiment 1.

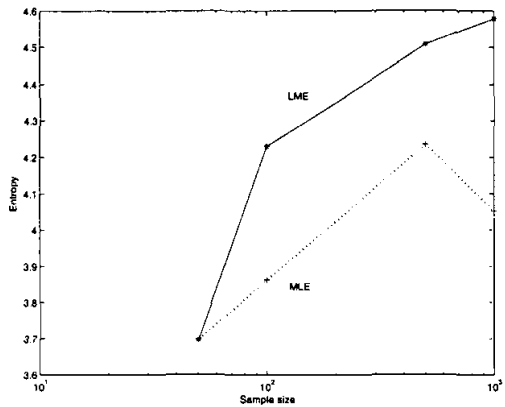

Figure 4: Average entropy of the MLE estimate versus the LME estimates in Experiment 1.

about the underlying model.

Figure 6 shows that LME obtained a significantly lower cross entropy than MLE.

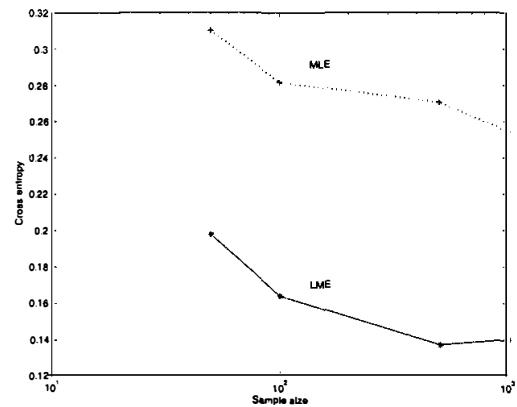

Figure 6: Average cross entropy between the true distribution and the MLE estimate versus the LME estimates in Experiment 2.

Figures 3 and 4 first show that the average log-likelihoods and average entropies of the models produced by LME and MLE respectively behave as expected. MLE clearly achieves higher log-likelihood than LME, however LME clearly produces models that have significantly higher entropy than MLE. The interesting outcome is that the two estimation strategies obtain significantly different cross entropies. Figure 5 reports the average cross entropy obtained by MLE and LME as a function of sample size, and shows the result that LME achieves substantially lower cross entropy than MLE. LME's advantage is especially pronounced at small sample sizes, and persists even when sample sizes as large as 1,000 are considered (Figure 5).

In our second experiment we used a generative model that was a Boltzmann machine with five observable and five hidden units. Specifically, we generated data with this architecture. The LME and MLE estimators still only inferred a Boltzmann machine with five observable and three hidden in this case, and hence were making an incorrect 'undercomplete' assumption

In our third experiment we used a generative model that was a Boltzmann machine with five observable and one hidden and the data were generated by this architecture. Again the LME and MLE estimators inferred Boltzmann machine with five observable and three hidden in this case, and hence were making an incorrect 'overcomplete' assumption about the underlying model.

Figure 7 shows that LME still obtained a significantly lower cross entropy than MLE.

Although these results are anecdotal, we have witnessed a similar outcome on several other models as well. Wider experimentation on synthetic data and real MRF application and theoretical analysis are necessary to confirm this as a general conclusion. More-over, it is worthwhile to discuss and compare with training by minimizing contrastive divergence (Hinton 2002).

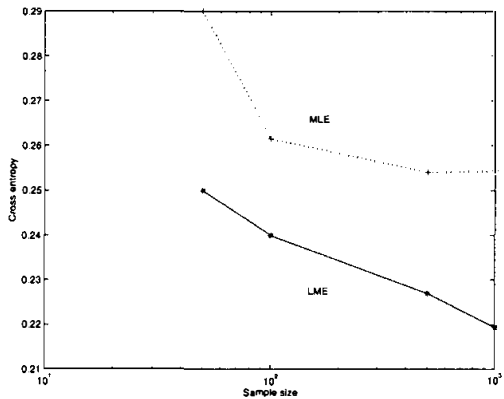

Figure 7: Average cross entropy between the true distribution and the MLE estimate versas the LME estimates in Experiment 3.

## 7　Conclusion

We have introduced a new inference technique, the latent maximum entropy principle, and used it to derive a new local parameter optimization method, EM-IS, and also a new selection technique, ME-EM-IS, for Boltzmann machines. These principles yield new training and estimation algorithms that perform very effectively when compared to the standard methods.

There remain several avenues to extend this work. More generally, by allowing binary-valued features of the form $f_V(w) = w_{v_1} w_{v_2} ... w_{v_n}$ for $v = (v_1,...v_n)$, a path in $G = (E, V)$, we construct models that are essentially "higher-order" Boltzmann machines (Sejnowski 1986).

Also, the training and estimation methods can still be improved by tackling them simultaneously. In this paper, by randomly choosing different starting points, we take the feasible log-linear model with maximum entropy value as the LME estimate. This procedure is computationally expensive. Thus it is worthwhile to develop an analogous deterministic annealing ME-EM-IS algorithm to automatically find feasible maximum entropy log-linear model for LME (Ueda and Nakano 1998).